\documentclass[sigconf]{acmart}

\copyrightyear{2026}
\acmYear{2026}
\setcopyright{cc}
\setcctype{by}
\acmConference[WWW '26]{Proceedings of the ACM Web Conference 2026}{April 13--17, 2026}{Dubai, United Arab Emirates}
\acmBooktitle{Proceedings of the ACM Web Conference 2026 (WWW '26), April 13--17, 2026, Dubai, United Arab Emirates}
\acmPrice{}
\acmDOI{10.1145/3774904.3792379}
\acmISBN{979-8-4007-2307-0/2026/04}

\AtBeginDocument{%
  }

\newtheorem{challenge}{Challenge}
\usepackage{amsmath,amsfonts}
\usepackage{balance}
\usepackage{algorithmic}
\usepackage{algorithm}
\usepackage{multirow}
\usepackage{subfigure}
\begin{document}

\title{Generalized Incremental Learning under Concept Drift across Evolving Data Streams}

\author{En Yu}
\affiliation{
  \institution{Australian Artificial Intelligence
Institute (AAII), \\University of Technology Sydney}
  \city{Sydney}
  \state{NSW}
  \country{Australia}
}
\email{en.yu-1@uts.edu.au}

\author{Jie Lu}
\authornote{Corresponding author.}
\affiliation{
  \institution{Australian Artificial Intelligence
Institute (AAII), \\University of Technology Sydney}
  \city{Sydney}
  \state{NSW}
  \country{Australia}
}
\email{jie.lu@uts.edu.au}

\author{Guangquan Zhang}
\affiliation{
  \institution{Australian Artificial Intelligence
Institute (AAII), \\University of Technology Sydney}
  \city{Sydney}
  \state{NSW}
  \country{Australia}
}
\email{guangquan.zhang@uts.edu.au}


\begin{abstract}
Real-world data streams exhibit inherent non-stationarity characterized by concept drift, posing significant challenges for adaptive learning systems. While existing methods address isolated distribution shifts, they overlook the critical co-evolution of label spaces and distributions under limited supervision and persistent uncertainty. To address this, we formalize Generalized Incremental Learning under Concept Drift (GILCD), characterizing the joint evolution of distributions and label spaces in open-environment streaming contexts, and propose a novel framework called Calibrated Source-Free Adaptation (CSFA). First, CSFA introduces a \emph{training-free prototype calibration} mechanism that dynamically fuses emerging prototypes with base representations, enabling stable new-class identification without optimization overhead.  Second, we design a novel source-free adaptation algorithm, i.e., \emph{Reliable Surrogate Gap Sharpness-aware} (RSGS) minimization. It integrates sharpness-aware perturbation loss optimization with surrogate gap minimization, while employing entropy-based uncertainty filtering to discard unreliable samples. This mechanism ensures robust distribution alignment and mitigates generalization degradation caused by uncertainties. Thus, CSFA establishes a unified framework for stable adaptation to evolving semantics and distributions in open-world streaming scenarios. Extensive experiments validate the superior performance and effectiveness of CSFA compared to SOTA approaches.
\end{abstract}




\maketitle

\section{Introduction}
\begin{figure}[t]
    \centering
    \includegraphics[width=0.95\columnwidth]{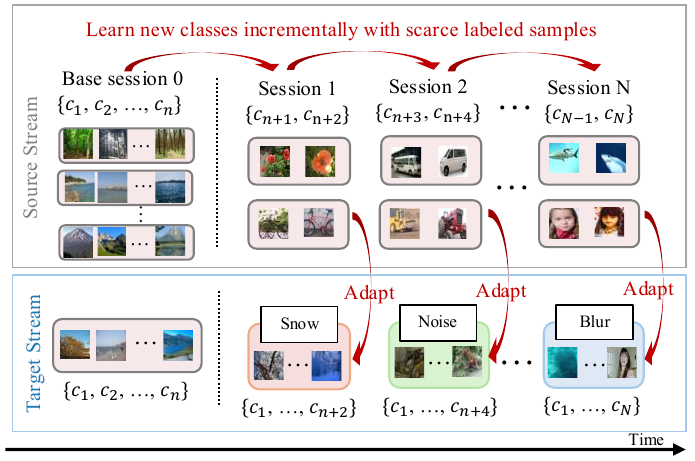}
    \caption{Illustration of our proposed Generalized Incremental Learning under Concept Drift (GILCD) setting.}
    \label{fig:setting}
\end{figure}

In machine learning, the conventional training process typically relies on pre-collected datasets. It assumes that training and test data ideally adhere to the same distribution, facilitating the effective generalization of trained models to test data. However, the World Wide Web, as a vast source of information, generates massive dynamic and continuously evolving data over time, known as data streams~\cite{gaudreault2024systematic,wang2025local}. 
These data streams are susceptible to changes in their underlying distribution, a phenomenon known as concept drift~\cite{lu2018learning}.
For instance, social media platforms continuously generate massive streams of user-uploaded images, where popular styles, filters, or meme formats evolve rapidly. Similarly, e-commerce and news sites constantly update product and event images, leading to distribution shifts that challenge models trained on static datasets.
In addition, these data streams cannot be stored for extended periods due to storage constraints or privacy concerns. Consequently, the model must be updated incrementally with limited newly arriving data while making fast adaptations for concept drift.


Previous research has demonstrated the effectiveness of concept drift adaptation techniques in managing changing data distributions. However, the majority of these methods are designed for individual streams with delayed labels, which limits their generalization in more complex drifting scenarios~\cite{song2021learning,yu2026drift}. For example, 
in a sophisticated e-commerce platform, the recommendation engine aims to provide highly personalized and timely suggestions to users. To achieve this, it simultaneously ingests and processes multiple data streams, each with distinct characteristics and data distributions~\cite{he2024invariant,zhang2023invariant}. Furthermore, although data collection is relatively straightforward, the labeling process involves significant time and labor costs, resulting in a hybrid scenario where numerous labeled and unlabeled streams arrive concurrently. To address this situation, multistream classification has been introduced, involving both labeled and unlabeled data streams with concept drifts~\cite{yu2024online}. This task aims to predict the labels of the target stream by transferring knowledge from one or multiple labeled source streams, while also handling the concept drift problem. However, these studies have overlooked a key issue in real-world applications: learning occurs continuously on incoming data streams, which may contain both data from new classes and new observations of old classes~\cite{he2020incremental,jiao2025otl}. For example, on an e-commerce platform, the recommendation system must learn to suggest products from newly introduced categories or brands while retaining its knowledge of a user's established preferences for existing ones.

While class-incremental learning has been developed to incorporate new semantics while mitigating catastrophic forgetting~\cite{zhou2024class}, it often presupposes the availability of abundant labeled data for supervised training. This reliance on extensive labeled data becomes particularly challenging in real-world data streams characterized by rapid data arrival and limited annotation resources. This has spurred interest in Few-Shot Class-Incremental Learning (FSCIL)~\cite{bai2020class}, which attempts to learn new classes from scarce labeled examples. Nevertheless, FSCIL typically assumes relatively stable data distributions, often overlooking the critical possibility that the newly arriving data itself might be subject to unforeseen distributional changes, i.e., concept drift. Consequently, a significant research gap emerges: existing approaches often falter when faced with the concurrent challenges of concept drift and the need to incrementally learn new classes with limited labeled data. They may either struggle to effectively incorporate novel classes or suffer substantial performance degradation when the new data exhibits distributional shifts~\cite{wang2024few,yang2025adapting}. This renders the model unable to accurately predict subsequent out-of-distribution samples, leading to a significant decline in overall performance.

To fill this research gap, we introduce Generalized Incremental Learning under Concept Drift (GILCD), a novel research setting. As shown in Figure~\ref{fig:setting}, GILCD models a scenario with two concurrent data streams (i.e., Source and Target streams) both evolving across sequential sessions over time. In incremental sessions, the source stream offers scarce labeled samples for newly emerging classes. Concurrently, the unlabeled target stream expands its label space to include all classes encountered so far. Critically, the target stream also undergoes concept drift with its data distribution shifting over time. The primary objective of GILCD is to learn new classes incrementally from the source stream while mitigating forgetting, and simultaneously adapting to the evolving distributions and expanding the class set of the target stream.

Therefore, we propose a \textbf{C}alibrated \textbf{S}ource-\textbf{F}ree \textbf{A}daptation (CSFA) framework, which addresses both the incremental learning of new classes from scarce data and the continuous adaptation to concept drift in a source-free manner. Specifically, we first pre-train a base model on the initial data-rich base session and subsequently freeze its feature extractor. This strategy provides a stable foundation for learning new classes and inherently mitigates catastrophic forgetting when new classes arrive. To incrementally incorporate novel classes using only the scarce labeled examples available in subsequent source stream sessions, CSFA employs an efficient \textit{training-free calibrated prototype strategy}. Instead of simply learning new class prototypes in isolation, which can be unreliable with limited data, this strategy (detailed in Section~\ref{sec:incremental_learning_detail}) leverages the well-learned base prototypes, and calibrates the biased novel prototypes by fusing them with a weighted combination of these established base prototypes. It thereby ensures more stable and generalizable representations for the new classes without requiring further optimization of the feature extractor.

Furthermore, CSFA addresses the persistent challenge of concept drift in the target stream through a novel source-free adaptation strategy called \textit{Reliable Surrogate Gap Sharpness-aware (RSGS) minimization}. Recognizing that direct access to source data is often infeasible during deployment, RSGS adapts the model using only the unlabeled target data. As elaborated in Section~\ref{sec:adaptation_detail}, RSGS goes beyond standard sharpness-aware minimization. It not only seeks a flat minimum by minimizing perturbation loss and the surrogate gap simultaneously, but critically it integrates a reliability-aware mechanism. This involves an indicator function to filter out high-entropy (i.e., uncertain or noisy) target samples from the adaptation process. This selective adaptation ensures robust distribution alignment between the perceived source knowledge and the evolving target stream, enhancing generalization to changing target distributions. The contributions can be summarized as,
\begin{itemize}
    \item  We introduce GILCD, a practical and challenging setting that formalizes the concurrent demands of class evolution and concept drift within evolving data streams. It offers a new direction for developing robust adaptive systems in real-world non-stationary environments.

    \item We propose a novel framework, CSFA, which integrates a \textit{training-free} prototype strategy for learning new classes from scarce labels and a robust \textit{source-free} mechanism for concept drift adaptation. CSFA also tackles label scarcity and data invisibility in evolving real-world scenarios, enabling efficient real-time decision-making.
    
     \item We introduce the RSGS minimization algorithm that enhances adaptation robustness by uniquely incorporating an entropy-based filter to discard high-uncertainty samples during the source-free process. Comprehensive experiments demonstrate the superiority of our method.
\end{itemize}

\begin{figure*}
    \centering
    \includegraphics[width=0.95\textwidth]{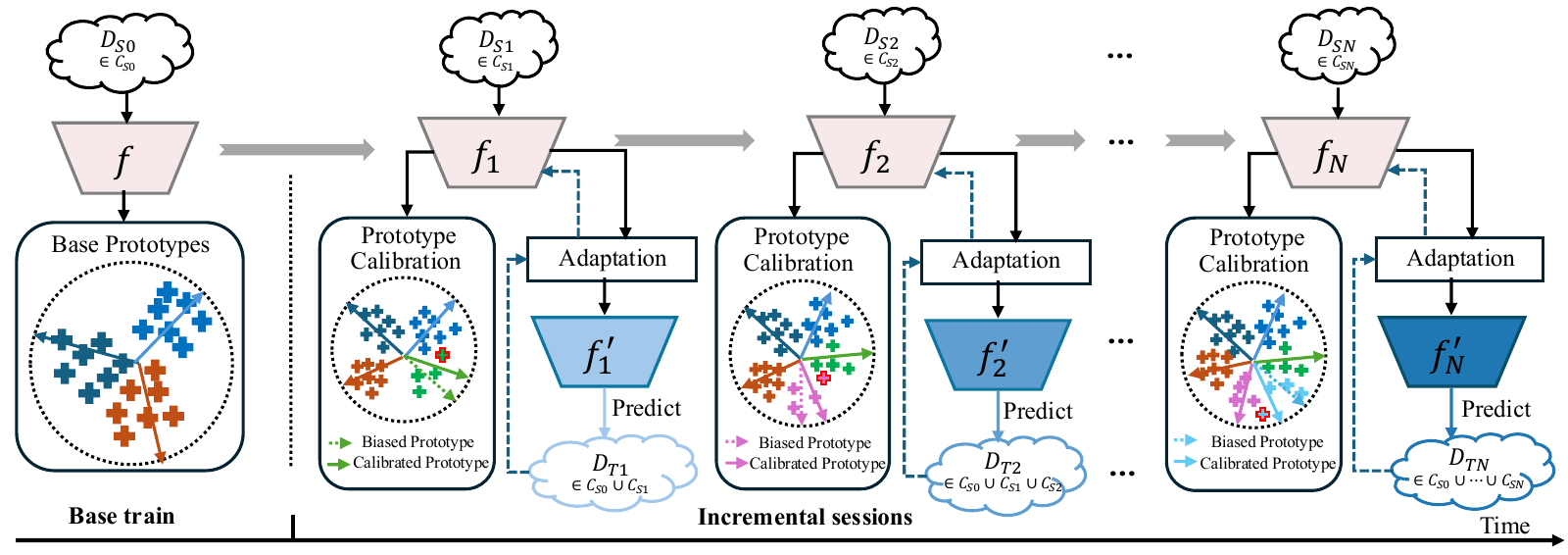}
    \caption{Illustration of the proposed CSFA framework. Firstly, a base model is trained on the large-scale base session ${\mathcal D}_{S0}$. When dealing with new sessions with limited samples, the pre-trained feature extractor is frozen, and a calibrated prototype-based strategy is adopted to learn novel prototypes incrementally. In addition, it further addresses the covariate shift and target drift by introducing a source-free adaptation strategy, which jointly minimizes the entropy and the sharpness of the entropy of those reliable target samples from the new distribution.}
    \label{fig:framework}
\end{figure*}

\section{Related Work}

\textbf{Data Stream Learning.}
With the rapid growth of the Web, massive streams of user-generated and dynamic content (e.g., social media images, news updates, product catalogs) are continuously produced in real time~\cite{li2025pontus,wang2025local,yu2025learning}. Unlike static datasets, these data streams are inherently non-stationary where the underlying distribution $P_t(X,y)$ often shifts to $P_{t+1}(X,y)$ over time--a phenomenon known as \textit{concept drift}~\cite{lu2018learning,yu2024gnn}. Such drift severely challenges the stability and generalization of deployed models, as classifiers trained on past data may quickly become outdated. To cope with these issues, prior work has explored adaptive strategies including window-based updates~\cite{xu2017dynamic}, instance selection~\cite{gomes2017adaptive}, and ensemble methods~\cite{wang2016online}. More recent advances extend to multistream scenarios, where knowledge must be transferred across heterogeneous or partially labeled streams~\cite{haque2017fusion,yu2022meta,yu2024fuzzy}. Nevertheless, existing approaches often assume a fixed label space, while in real-world Web environments new classes emerge continuously, requiring algorithms to jointly handle both concept drift and class evolution.

\noindent
\textbf{Class-Incremental Learning (CIL) and Test-Time Adaptation (TTA).}
CIL aims to build classifiers over an expanding label space while mitigating catastrophic forgetting~\cite{rebuffi2017icarl}. Approaches include replay, knowledge distillation, and model expansion~\cite{gu2022not,wang2024few}. Few-shot CIL (FSCIL) addresses data scarcity by updating with limited labels~\cite{tao2020few,lv2025grasp}. Nonetheless, most CIL and FSCIL studies assume static distributions, leaving distribution shifts unaddressed. TTA methods adapt models during inference using self-supervised learning or entropy minimization~\cite{niu2022towards,chen2022contrastive}. Source-free domain adaptation similarly operates without access to source data~\cite{liang2020we}. These methods enhance distributional robustness but ignore class evolution. 


\section{Methodology}
\label{methodology}

\subsection{Preliminary}
\label{sec:preliminary}
As shown in Figure~\ref{fig:setting}, we assume there are two data streams in the GILCD setting and each stream contains $N+1$ sessions, including a base session and $N$ incremental sessions. We represent a sequence of disjoint source sessions by ${\mathcal D}_S=\{{\mathcal D}_{S0},{\mathcal D}_{S1},...,{\mathcal D}_{Sn}\}$, where ${\mathcal D}_{S0}$ is a large-scale base dataset with abundant labeled samples and the following ${\mathcal D}_{Si}, i>0$ are all novel sessions with scarce labeled samples. For the source data ${\mathcal D}_{Si}$ in the $i$-th session, we further define it as $\{(x_{i},y_{i})\}_{i=1}^{|N_{i}|}$ with the corresponding label space ${\mathcal C}_{Si}$, where ${\mathcal C}_{Si} \cap {\mathcal C}_{Sj} = \emptyset$. 
Accordingly, an evolving target stream is defined as ${\mathcal D}_T=\{{\mathcal D}_{T0},{\mathcal D}_{T1},...,{\mathcal D}_{Tn}\}$.
The target label space and distribution of session $i$ are denoted as ${\mathcal C}_{Ti}$ and ${\mathcal P}_{Ti}$, respectively. Specifically, the target label space of the $i$-th session contains all seen classes during inference. Furthermore, the distributions of target sessions also change over time, i.e., ${\mathcal P}_{Ti} \neq {\mathcal P}_{Tj}, i \neq j$. Overall, all challenges can be summarized as follows,
\begin{challenge}[{Class Incremental}]
  For the source stream ${\mathcal D}_S$, the corresponding label space ${\mathcal C}_{Si}$ of each session ${\mathcal D}_{Si}, i>0$ incrementally evolves, i.e., ${\mathcal C}_{Si} \cap {\mathcal C}_{Sj} = \emptyset$, whereas the target label space of the $i$-th session contains all seen classes during inference, i.e., $\mathcal{C}_{Ti}=\bigcup_{j=0}^{i}\mathcal{C}_{Sj}$. 
\end{challenge}
\begin{challenge}[{Scarcity of Labels}]
   Only limited labeled samples are provided to the source streams ${\mathcal D}_S=\{{\mathcal D}_{S0},{\mathcal D}_{S1},...,{\mathcal D}_{Sn}\}$, leaving the target stream entirely unlabeled ${\mathcal D}_T=\{{\mathcal D}_{T0},{\mathcal D}_{T1},...,{\mathcal D}_{Tn}\}$. A session in the source stream can alternatively be described as an $N$-way $K$-shot classification task, which involves $N$ classes and $K$ labeled examples for each class. Consequently, the challenge lies in achieving accurate predictions in the target stream, where no labeled samples are available.
\end{challenge}
\begin{challenge}[{Covariate Shift}]
      Denoting ${\mathcal P}_{Si}$ and ${\mathcal P}_{Ti}$ as the distributions of ${\mathcal D}_{Si}$ and ${\mathcal D}_{Ti}$ in the incremental sessions, all streams at the same session are related but exhibit covariate shift ${\mathcal P}_{Si}(\boldsymbol{x})  \neq{\mathcal P}_{Ti}( \boldsymbol{x})$
\end{challenge}
\begin{challenge}[{Target Drift}]
     Target drift refers to changes in the distributions of target sessions over time, i.e., ${\mathcal P}_{Ti} \neq {\mathcal P}_{Tj}, i \neq j$.
\end{challenge}

The primary objective of GILCD is to develop a unified classification model $f(\mathbf{x})$ capable of effectively handling all these challenges. This task requires the model to achieve three key goals: \textit{1) incrementally learn new classes with limited labeled data without forgetting old classes}; \textit{2) continuously make adaptations from the labeled source stream to the unlabeled target stream}; and \textit{3) dynamically adapt to new distributions for target stream prediction}.

\subsection{Overview of CSFA}

To address the inherent challenges of the GILCD task, we propose a novel method called CSFA, illustrated in Figure~\ref{fig:framework}. Initially, in response to \emph{Challenges 1 and 2}, we commence by pre-training a base model on abundant labeled data from the base session. Subsequently, we freeze this model and utilize it as a feature extractor when encountering new classes, thus mitigating the adverse effects of catastrophic forgetting. We introduce a simple yet effective calibrated prototype-based strategy to overcome the challenge of limited data availability during incremental learning. This strategy not only examines the prototypes of weighted base prototypes but also integrates biased prototypes of new classes with base prototypes to determine the corresponding classifier weights (see Section~\ref{sec:incremental_learning_detail}). Importantly, this approach eliminates the need for additional optimization procedures after the base training.

Furthermore, we address \emph{Challenges 3 and 4} by devising a source-free adaptation strategy. This strategy operates by simultaneously minimizing the perturbation loss and the surrogate gap as well as integrating a reliable indicator function to filter out samples with high entropy. RSGS not only removes the high loss within a neighborhood but also ensures that the obtained minimum is situated within a flat region. A detailed analysis is elucidated in Section~\ref{sec:adaptation_detail}. Through this multifaceted approach, we aim to fortify the robustness and adaptability of incremental learning systems, ensuring their efficacy in dynamic environments.

\subsection{Training-free Class-Incremental Learning}
\label{sec:incremental_learning_detail}
\noindent
\textbf{Base model training:} 
Firstly, we use the collected data ${\mathcal D}_{S0} \in \{(x_{i},y_{i})\}_{i=1}^{|N_{0}|}$ with abundant labeled training data to train a base model $f(\mathbf{x})$, which follows a standard classification pipeline and is optimized by cross-entropy loss,
\begin{equation}
\centering
\sum_{(\mathbf{x}_j,y_j) \in {\mathcal D}_{S0}} \mathcal{L}_{E}\left(f\left(\mathbf{x}_j; \mathcal{\theta}\right),y_j\right)
\end{equation}
where $ \mathcal{L}_{E}$ denotes the cross-entropy loss. $x_{i} \in \mathbb{R}^{d}$ is a training sample with label $y_{i}$ from the base label space. Following \cite{rebuffi2017icarl,wang2022learning}, the model can be denoted as $f(\mathbf{x})=W^{\top}\phi(\mathbf{x})$, where $\phi(\cdot):\mathbb{R}^{D}\rightarrow\mathbb{R}^{d}$ is the feature extractor and $W\in\mathbb{R}^{d\times N}$ is the classification head. If there are $N$ classes, the classifier head can be denoted by the $N$ prototypes, i.e., $W=[c_{1},c_{2},\ldots,c_{N}]$.

\noindent
\textbf{Incremental learning:}
Prototype classifier~\cite{snell2017prototypical} is a popular approach employed in few-shot learning tasks. To achieve this, the feature extractor $\phi(\cdot)$ trained by the base session remains fixed to minimize catastrophic forgetting, and then we utilize the mean feature $\mathbf{c}_k$ as the prototype to capture the most common pattern observed in each class,
\begin{equation}
\mathbf{c}_k=\frac{1}{{num}_k}\sum_{y_{i}=k}f_\phi(\mathbf{x}_i),
\label{protonet}
\end{equation}
where ${num}_k$ is the number of samples of the $k$th class. The prototypes of new classes are utilized as the corresponding classifier weights~\cite{zhang2021few,zhou2022forward}. Then we can estimate the probability of each class $k$ via the dot products between the features and each prototype,
\begin{equation}
p\left(y_i=k|\phi(\mathbf{x}_i)\right) = \frac{\exp(\phi(\mathbf{x}_i)\cdot\boldsymbol{c}_k)}{\sum_j\exp(\phi(\mathbf{x}_i)\cdot\boldsymbol{c}_j)}.
\end{equation}

\begin{figure}[htbp]
    \centering
    \includegraphics[width=\columnwidth]{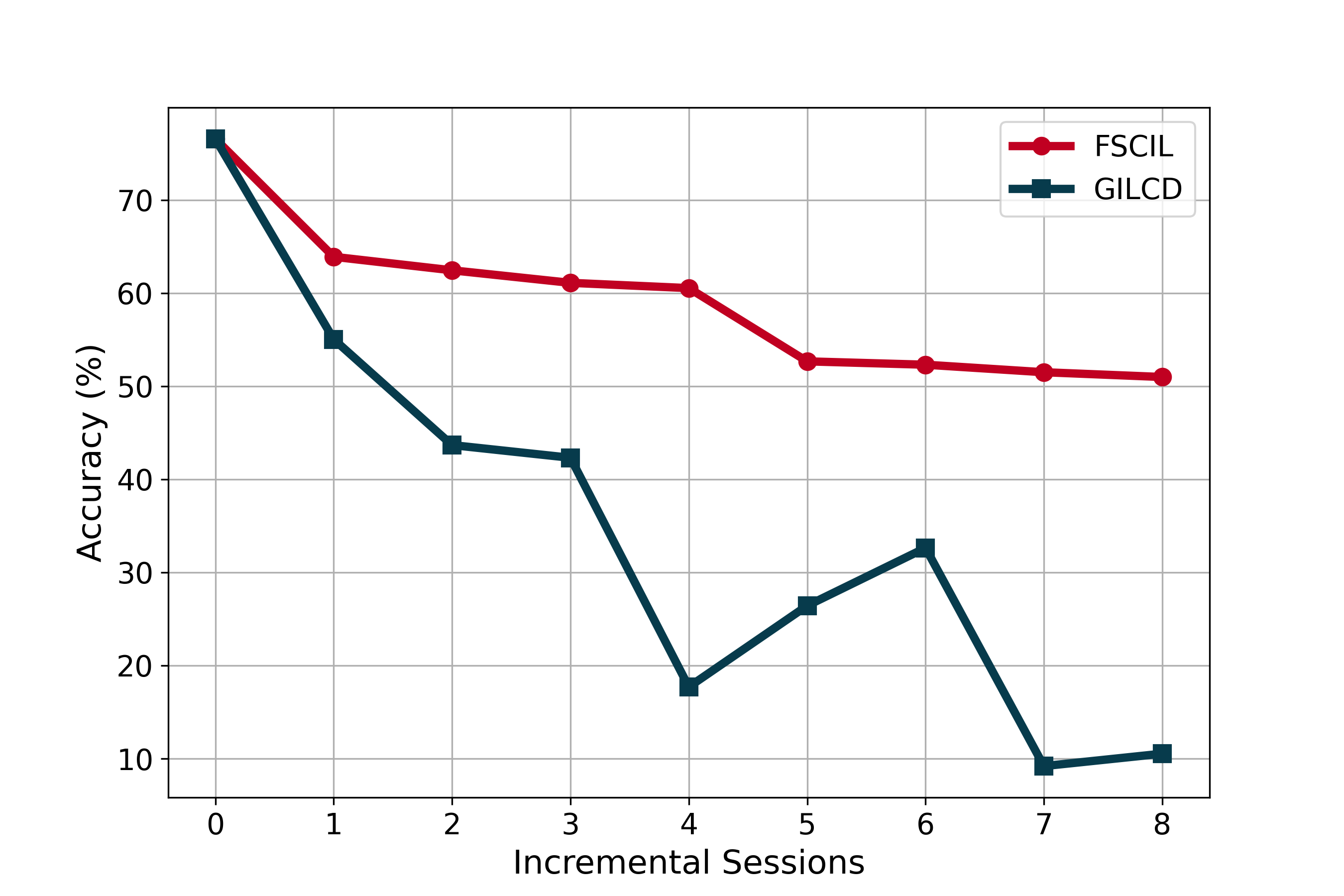}
    \caption{Comparison of TEEN's performance between FSCIL and GILCD settings. We compare the performance of TEEN \cite{wang2020tent} in the stationary FSCIL scenario (i.e., CIFAR100) with our proposed GILCD setting with concept drift (i.e., CIFAR100-C). It is evident that the current FSCIL methods fail to address the distribution shift in the GILCD scenario.}
    \label{fig:tenn_drops}
\end{figure}

However, in the GILCD setting, the number of samples in the new class is extremely limited, leading to severe bias in the empirical prototypes of the novel classes. Thus, we utilize the well-learned base prototypes to calibrate the biased prototypes of novel classes during incremental sessions~\cite{wang2024few}. To achieve this, we define the base session as comprising $n$ classes and the incremental session as containing $C$ additional classes. Hence, the base prototypes and new prototypes are denoted as $c_{b}(1 \leq b \leq n)$ and $c_{new}(n < b \leq n+C)$, respectively. Since the base session includes adequate samples in every class, the model can effectively capture the distribution of base classes and derive guaranteed prototypes. Consequently, we can leverage these base prototypes $c_{b}$ to calibrate the prototypes for newly coming classes $c_{new}$ by,
\begin{equation}
\bar{c}_{new}=\alpha c_n+(1-\alpha)\Delta c_{new},
\end{equation}
where $\bar{c}_{new}$ is the calibrated new prototype, and hyperparameter $\alpha$ controls the calibration strength of new prototypes. $\Delta c_{new}$ denotes the calibration item, which is constructed by weighted base prototypes. Specifically, the cosine similarity $S_{b,new}$ between $c_{new}$ and a base prototype $c_{b}$ is obtained by, 
\begin{equation}
S_{b,new}=\frac{c_b\cdot c_{new}}{\|c_b\|\cdot\|c_{new}\|}\cdot\tau, 
\end{equation}
where $\tau$ is the scaling hyperparameter and $\tau > 0$. Then the weight of the new class prototype $c_{new}$ can be represented as the softmax output over all base prototypes,
\begin{equation}
w_{b,new}=\frac{e^{S_{b,new}}}{\sum_{i=1}^{n}e^{S_{i,new}}},
\label{new_weight}
\end{equation}
Finally, the well-calibrated prototypes of novel classes are defined as,
\begin{equation}
\begin{aligned}
  \bar{c}_{new}&=\alpha c_{new}+(1-\alpha)\Delta c_{new} \\ 
  &=\alpha c_{new}+(1-\alpha)\sum\limits_{b=1}^{n}w_{b,new}c_b. 
\end{aligned}
\label{calibrated_proto}
\end{equation}

\begin{algorithm}[t!]
\caption{The pipeline of proposed CSFA}
\label{alg1}
\renewcommand{\algorithmicrequire}{\textbf{Input:}}
\renewcommand{\algorithmicensure}{\textbf{Output:}}
\begin{algorithmic}[1]
    \REQUIRE Source stream ${\mathcal D}_S=\{{\mathcal D}_{S0},{\mathcal D}_{S1},...,{\mathcal D}_{Sn}\}$; Target stream ${\mathcal D}_T=\{{\mathcal D}_{T0},{\mathcal D}_{T1},...,{\mathcal D}_{Tn}\}$, and model $f$.
    \ENSURE Predictions for target stream.
    \STATE \textbf{Base train:} 
    \STATE Train a base model using ${\mathcal D}_{S0}$. 
    \STATE Calculate the base prototypes via Eq.~\eqref{protonet}
    \STATE \textbf{Incremental learning:}
    \FOR{$i$ = 1 : $n$}
        \STATE Calculate the weight of new prototypes via Eq.~\eqref{new_weight}.
        \STATE Calculate the calibrated prototypes via Eq.~\eqref{calibrated_proto}.
        \STATE Adapt parameters by minimizing Eq.~\eqref{eq:RSGS}.
    \ENDFOR
\end{algorithmic}
\end{algorithm}

\subsection{Source-Free Drift Adaptation}
\label{sec:adaptation_detail}
Beyond the challenges posed by class incremental, the GILCD setting is severely impacted by: 1) significant covariate shift between the source and target streams (i.e., \emph{Challenge 3}), and 2) unceasing concept drift within the target stream (i.e., \emph{Challenge 4}). These concurrent distributional shifts can cause drastic performance degradation in learned models. As shown in Figure~\ref{fig:tenn_drops}, we compare the performance of the classic FSCIL method TEEN~\cite{wang2020tent} in a stable scenario with our proposed GILCD
setting with distribution changes. Existing FSCIL approaches are largely unprepared for such profound distributional dynamics. This necessitates a robust continuous adaptation mechanism for the unlabeled evolving target stream. This need is further exacerbated by the source stream's limited labeled samples and the entirely unlabeled target stream, which severely complicates knowledge transfer.

Consequently, our primary goal during adaptation is to enhance the model's generalization to the changing distributions within the target stream under source-free conditions. However, naively minimizing empirical risk (e.g., prediction entropy) on incoming unlabeled target samples can lead to overfitting to transient patterns or convergence to a sharp minima in the loss landscape, resulting in poor out-of-distribution performance and potential model collapse~\cite{li2023robustness, niu2022towards}. Therefore, an effective adaptation strategy should not only seek solutions in flat regions of the loss surface~\cite{foret2020sharpness}, but also ensure that the adaptation process itself is reliable, especially when dealing with real-world uncertain data streams.

\begin{figure}[!t]
    \centering
    \includegraphics[width=0.85\columnwidth]{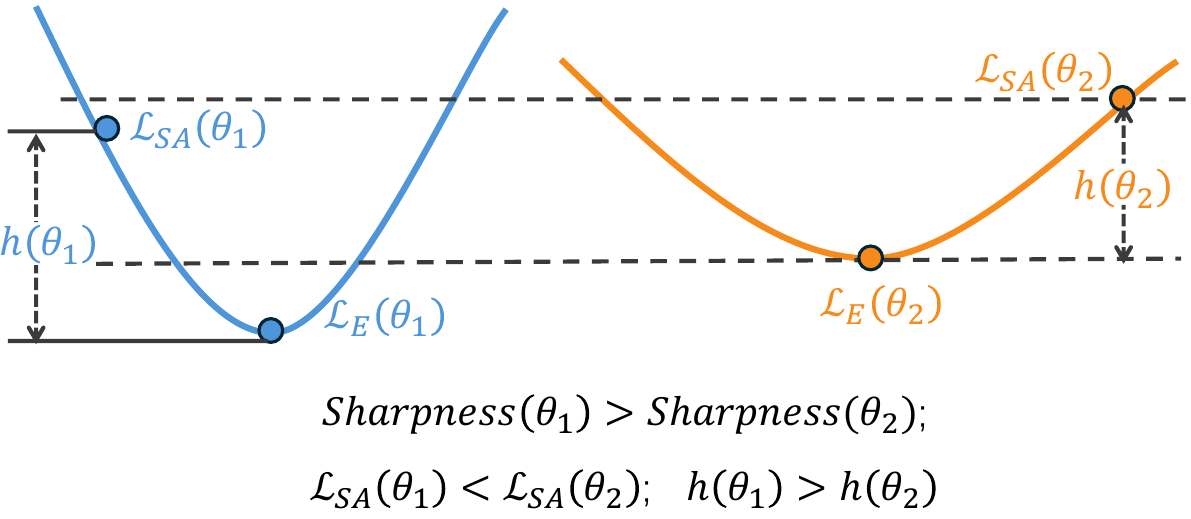}
    \caption{Consider a scenario with a sharp local minimum $\theta_{1}$ and a flat local minimum $\theta_{2}$. The loss surface around $\theta_{2}$ appears flatter than that around $\theta_{1}$. However, SAM exhibits bias towards selecting $\theta_{1}$ over $\theta_{2}$ because $\mathcal{L}_{SA}(\theta_{1}) < \mathcal{L}_{SA}(\theta_{2})$. Instead, the surrogate gap $h(\theta)$ provides a better description of the sharpness of the loss surface. A smaller $h(\theta_{2})$ indicates that $\theta_{2}$ is flatter than $\theta_{1}$.}
    \label{fig:sharpness_curve}
\end{figure}

To achieve this, we propose \textbf{R}eliable \textbf{S}urrogate \textbf{G}ap \textbf{S}harpness-aware (RSGS) minimization. The core idea behind RSGS is twofold:
\textit{1) Promoting Flat Minima for Generalization:} To encourage convergence to flat regions, RSGS aims to minimize not just the empirical loss $\mathcal{L}_{E}(\mathcal{D};\theta)$ on target data $\mathcal{D}$, but also the sharpness of the loss landscape. This involves considering the worst-case loss $\mathcal{L}_{SA}({\mathcal D} ; \theta) \triangleq \max _{\|\epsilon\|_2 \leq \rho} \mathcal{L}_{E}({\mathcal D} ; \theta+\epsilon)$ within a neighborhood $\rho$ of the current parameters $\theta$, and the surrogate gap $h(\theta)\triangleq \mathcal{L}_{SA}({\mathcal D}; \theta)-\mathcal{L}_{E}({\mathcal D}; \theta)$, which quantifies this sharpness \cite{zhuang2022surrogate}. Minimizing both $\mathcal{L}_{SA}$ and $h(\theta)$ steers the model towards flatter solutions, as illustrated conceptually in Figure \ref{fig:sharpness_curve}, where a flatter $\theta_2$ typically exhibits a smaller $h(\theta_2)$.
\textit{2) Ensuring Reliable Adaptation with Uncertain Samples:} Unlabeled target samples in evolving streams can be noisy or uncertain~\cite{zhuang2022surrogate}. Blindly adapting to all such samples can introduce substantial gradients, negatively impacting stability and leading to model collapse. RSGS addresses it by incorporating a reliable filter mechanism. Thus, the RSGS objective is defined as:
\begin{equation}
    \mathcal{L}_{RSGS} = \min _{\theta}( G(\mathcal{D})\mathcal{L}_{SA}(\mathcal{D}; \theta), h(\theta)),
    \label{eq:RSGS}
\end{equation}
where $G(\mathcal{D})\triangleq\mathbb{I}_{\{\mathcal{L}_{E}(\mathcal{D};\theta)<E_0\}}(\mathcal{D})$ is the reliable indicator function. This function actively filters out samples with high entropy, i.e., $\mathcal{L}_{E}(\mathcal{D};\theta) \ge E_0$, where $E_0$ is a predefined confidence threshold, set to $0.4 \times \ln 1000$ in this work. This effectively removes low-confidence or high-uncertainty samples from the sharpness-aware optimization process~\cite{niu2022towards}.

\begin{table*}[htbp]
  \centering
  \caption{Comparison with CIL/FSL/FSCIL baselines on CUB200-C dataset.
   The corruptions for sessions 1-10 are 'pixelate',  'elastic transform', 'defocus blur', 'zoom blur', 'glass blur', 'snow', 'fog', 'Gaussian noise', 'shot noise', 'impulse noise'.}
    \begin{tabular}{ccccccccccccc}
    \toprule
    \multirow{2}[4]{*}{Method} & \multicolumn{11}{c}{Accuracy in each session (\%)}                                    & \multirow{2}[4]{*}{Average Accuracy (\%)} \\
\cmidrule{2-12}          & 0     & 1     & 2     & 3     & 4     & 5     & 6     & 7     & 8     & 9     & 10    &  \\
    \midrule
    iCaRL & 76.73 & 56.25 & 50.21 & 41.79 & 23.98 & 20.42 & 10.76 & 15.49 & 10.04 & 9.79  & 9.42  & 29.53 \\
    ProtoNet & 77.81 & 56.97 & 53.42 & 44.32 & 25.01 & 23.18 & 10.91 & 17.62 & 10.71 & 11.01 & 9.37  & 30.94 \\
    TOPIC & 77.49 & 58.32 & 55.14 & 45.01 & 25.32 & 24.19 & 12.11 & 19.73 & 10.91 & 11.57 & 9.09  & 31.72 \\
    CEC   & 79.62 & 60.79 & 57.48 & 48.32 & 27.03 & 26.21 & 14.37 & 22.79 & 12.35 & 13.2  & 10.17 & 33.85 \\
    FACT  & 80.73 & 61.01 & 58.76 & 47.40  & 28.63 & 27.42 & 15.48 & 23.21 & 13.72 & 14.37 & 10.09 & 34.62 \\
    TEEN  & 80.29 & 62.39 & 61.31 & 47.59 & 28.40  & 26.77 & 16.02 & 24.10  & 13.39 & 15.54 & 11.10  & 37.58 \\
    \textbf{Ours} & \textbf{81.87} & \textbf{65.23} & \textbf{62.25} & \textbf{49.21} & \textbf{28.77} & \textbf{34.23} & \textbf{33.96} & \textbf{36.37} & \textbf{26.61} & \textbf{27.53} & \textbf{20.83} & \textbf{42.44} \\
    \bottomrule
    \end{tabular}%
  \label{tab:CUB200-C}%
\end{table*}

\begin{table*}[htbp]
  \centering
  \caption{Comparison with CIL/FSL/FSCIL baselines on CIFAR100-C and miniImageNet-C datasets. The corruptions for sessions 1-8 are 'contrast', 'elastic transform', 'zoom blur', 'glass blur', 'frost', 'fog', 'Gaussian noise', 'shot noise'.}
    \begin{tabular}{cccccccccccc}
    \toprule
    \multirow{2}[4]{*}{Datasets} & \multirow{2}[4]{*}{Methods} & \multicolumn{9}{c}{Accuracy in each session (\%)}                     & \multirow{2}[4]{*}{Average Accuracy (\%)} \\
\cmidrule{3-11}          &       & 0     & 1     & 2     & 3     & 4     & 5     & 6     & 7     & 8     &  \\
    \midrule
    \multirow{7}[2]{*}{CIFAR 100-C} & iCaRL & 69.42 & 50.67 & 37.85 & 34.31 & 14.87 & 20.98 & 25.69 & 10.04 & 8.37  & 30.24 \\
          & ProtoNet & 72.31 & 51.49 & 38.72 & 35.07 & 14.98 & 22.04 & 27.72 & 9.87  & 9.21  & 31.27 \\
          & TOPIC & 70.42 & 50.77 & 37.83 & 36.01 & 14.03 & 21.42 & 27.56 & 9.52  & 8.77  & 30.70 \\
          & CEC   & 75.91 & 53.17 & 39.43 & 38.74 & 16.53 & 23.92 & 30.19 & 9.37  & 8.91  & 32.91 \\
          & FACT  & 76.03 & 54.21 & 40.37 & 40.01 & 17.82 & 25.73 & 33.71 & 9.24  & 9.76  & 34.10 \\
          & TEEN  & 76.55 & 55.03 & \textbf{43.67} & 42.31 & 17.70  & 26.42 & 32.63 & 9.20   & 10.53 & 34.89 \\
          & \textbf{Ours} & \textbf{76.55} & \textbf{55.23} & 41.02 & \textbf{46.56} & \textbf{25.94} & \textbf{34.44} & \textbf{35.23} & \textbf{21.81} & \textbf{20.35} & \textbf{39.68} \\
    \midrule
    \midrule
    \multirow{7}[2]{*}{miniImageNet-C} & iCaRL & 71.68 & 10.04 & 41.73 & 25.91 & 10.44 & 25.61 & 32.33 & 10.01 & 14.17 & 26.88 \\
          & ProtoNet & 73.92 & 11.09 & 44.27 & 26.29 & 10.23 & 24.36 & 33.08 & 10.44 & 14.32 & 27.56 \\
          & TOPIC & 74.08 & 10.71 & 46.33 & 30.21 & 11.84 & 28.17 & 35.14 & 13.25 & 15.47 & 29.47 \\
          & CEC   & 73.76 & 11.42 & 47.18 & 32.94 & 13.51 & 29.57 & 36.25 & 13.93 & 17.19 & 30.64 \\
          & FACT  & 75.21 & 11.78 & 49.52 & 28.49 & 11.87 & 30.21 & 38.64 & 15.17 & 18.03 & 30.99 \\
          & TEEN  & 75.36 & 11.60  & 49.81 & 27.65 & 12.02 & 32.01 & 39.99 & 15.06 & 16.85 & 31.15 \\
          & \textbf{Ours} & \textbf{75.37} & \textbf{13.51} & \textbf{55.71} & \textbf{44.03} & \textbf{22.93} & \textbf{44.94} & \textbf{42.78} & \textbf{25.65} & \textbf{26.04} & \textbf{39.00} \\
    \bottomrule
    \end{tabular}%
  \label{tab:cifarAndImage}%
\end{table*}%

\noindent
\textbf{Optimization.}
The optimization of $\mathcal{L}_{RSGS}$ follows a two-step approach. Firstly, we employ gradient descent $\nabla G(\mathcal{D})\mathcal{L}_{SA}(\mathcal{D}; \theta)$ to minimize the reliable sharpness-aware entropy loss $G(\mathcal{D})\mathcal{L}_{SA}(\mathcal{D}; \theta)$.  Specifically, the sharpness $\mathcal{L}_{SA}$ is quantified by the maximal change of entropy between $\theta$ and $\theta+\epsilon$. To address this problem, we use the first-order Taylor expansion to approximate its solution by,
\begin{equation}
\begin{aligned}
   \epsilon^*(\theta) &\triangleq \underset{\|\epsilon\|_2 \leq \rho}{\arg \max } \mathcal{L}_{E}({\mathcal D} ; \theta+\epsilon) \\
   &\approx \underset{\|\epsilon\|_2 \leq \rho}{\arg \max } \mathcal{L}_{E}({\mathcal D} ; \theta)+\epsilon^T \nabla \mathcal{L}_{E}({\mathcal D} ; \theta)\\
   &=\underset{\|\epsilon\|_2 \leq \rho}{\arg \max } \epsilon^T \nabla_{\theta} \mathcal{L}_{E}({\mathcal D} ; \theta) . 
\end{aligned}
\end{equation}
Subsequently, the solution to this approximation, denoted as $\hat{\epsilon}(\theta)$, is derived from resolving a classical dual norm problem,
\begin{equation}
\begin{aligned}
    &\hat{\epsilon}(\theta)
    =\rho \operatorname{sign}\left(\nabla \mathcal{L}_{E}({\mathcal D} ; \theta)\right) \frac{\left|\nabla \mathcal{L}_{E}({\mathcal D} ; \theta)\right|}{\left\|\nabla \mathcal{L}_{E}({\mathcal D} ; \theta)\right\|_2}.
\end{aligned}
\end{equation}
Substituting $\hat{\boldsymbol{\epsilon}}(\Theta)$ into $\mathcal{L}_{SA}$ and omitting second-order terms to expedite computation yields the final gradient approximation,
\begin{equation}
\left.\nabla \mathcal{L}_{SA}({\mathcal D} ; \theta) \approx \nabla \mathcal{L}_{E}({\mathcal D} ; \theta)\right|_{\theta+\hat{\epsilon}(\theta)},
\end{equation}
Additionally, to prevent model collapse from extremely low entropy values during adaptation, we employ a moving average $e_m$ (with decay 0.9) of the entropy loss $\mathcal{L}_{E}$. If $e_m$ drops below a threshold $e_0$ (set to 0.2), the parameters $\theta$ are reset to their state at the beginning of the current adaptation phase, promoting stability.

Secondly, we decompose the gradient $\nabla G(\mathcal{D})\mathcal{L}_{E}(\mathcal{D}; \theta)$ of the reliable entropy loss $G(\mathcal{D})\mathcal{L}_{E}(\mathcal{D}; \theta)$ into two components that are parallel and orthogonal to $\nabla G(\mathcal{D})\mathcal{L}_{SA}(\mathcal{D}; \theta)$, i.e., $\nabla G(\mathcal{D})\mathcal{L}_{E}(\mathcal{D}; \theta)_{\parallel}$ and $\nabla G(\mathcal{D})\mathcal{L}_{E}(\theta;\mathcal{D})_{\perp}$. Subsequently, it performs an ascent step in $\nabla G(\mathcal{D})\mathcal{L}_{E}(\theta;\mathcal{D})_{\perp}$ to minimize the surrogate gap $h(\theta)$. Thus the final gradient direction of RSGS can be formulated as,
\begin{equation}
\nabla\mathcal{L}_{RSGS}=\nabla G(\mathcal{D})\mathcal{L}_{SA}(\mathcal{D}; \theta)-\beta\nabla G(\mathcal{D})\mathcal{L}_{E}(\mathcal{D}; \theta)_{\perp}.
\end{equation}
where $\beta$ is a hyperparameter controlling the ascent step size. 

In summary, our proposed RSGS is a unique sharpness-aware minimization strategy for improved generalization, equipped with a crucial reliability filter that discards uncertain high-entropy target samples. This dual mechanism ensures that the model converges to a flat and robust minimum, effectively aligning distributions and generalizing well to changing target streams.

\section{Experiments}
\label{experiments}
\subsection{Benchmarks}
In our experiments, we employed three commonly used datasets for few-shot class incremental learning, i.e., CIFAR100 \cite{krizhevsky2009learning}, CUB200-2011 \cite{wah2011caltech}, and miniImageNet \cite{russakovsky2015imagenet}. 

\noindent
\textbf{Source stream:} Following the FSCIL paradigm~\cite{bai2020class,tao2020few}, we separate the training sets of CIFAR-100, miniImageNet, and CUB200-2011 into base and incremental sessions to construct the source stream. Specifically, for CIFAR-100 and miniImageNet, the training data is initially partitioned into 60 base classes, with the remaining 40 classes divided into eight 5-way 5-shot few-shot classification tasks. As for CUB200, the 200 classes are divided into 100 base classes and 100 incremental classes. The new incremental classes are then structured into 10-way 5-shot incremental tasks. 

\noindent
\textbf{Target stream:} To simulate various distributions, we synthesize three datasets CIFAR100-C, CUB200-C, and miniImageNet-C by introducing visual corruptions~\cite{hendrycks2018benchmarking}. Specifically, we added 15 types of corruption across four main categories (noise, blur, weather, and digital) with severity levels 5. Then, we randomly selected one type of corruption from the 15 available types for each incremental session, ensuring that the target stream exhibits a covariate shift compared to the source stream. Furthermore, it introduces distribution drift in the target stream over time.

\subsection{Baselines}
To validate our method, we compare it with the state-of-the-art methods of CIL, FSL, and FSCIL, including: iCaRL~\cite{rebuffi2017icarl}, ProtoNet~\cite{snell2017prototypical}, TOPIC~\cite{tao2020few}, CEC~\cite{zhang2021few},FACT~\cite{zhou2022forward}, TEEN~\cite{wang2024few}. In addition, we also compare our proposed adaptation method with current TTA and Domain Generalization (DG)  methods to demonstrate its adaptability to different distributions: TENT~\cite{wang2020tent}, SAM~\cite{foret2020sharpness}, GSAM~\cite{zhuang2022surrogate} and SAR~\cite{niu2022towards}. Detailed introduction of each method and experiment implementations can be found in Appendix~\ref{app:baselines} and ~\ref{app:imple}.

\begin{table*}[htbp]
  \centering
  \caption{Comparison with TTA/DG baselines on CIFAR100-C and miniImageNet-C datasets. The corruptions for sessions 1-8 are 'contrast', 'elastic transform', 'zoom blur', 'glass blur', 'frost', 'fog', 'Gaussian noise', 'shot noise'.}
    \begin{tabular}{cccccccccccc}
    \toprule
    \multirow{2}[4]{*}{Datasets} & \multirow{2}[4]{*}{Methods} & \multicolumn{9}{c}{Accuracy in each session (\%)}                     & \multirow{2}[4]{*}{Average Accuracy (\%)} \\
\cmidrule{3-11}          &       & 0     & 1     & 2     & 3     & 4     & 5     & 6     & 7     & 8     &  \\
    \midrule
    \multirow{5}[2]{*}{CIFAR100-C } & +TENT & 76.55 & 53.62 & 38.79 & 43.21 & 23.62 & 30.87 & 32.59 & 18.76 & 18.06 & 37.34 \\
          & +SAM  & 76.55 & 54.57 & 39.51 & 45.18 & 24.42 & 32.19 & 34.01 & 19.73 & 18.23 & 38.27 \\
          & +GSAM & 76.55 & 55.07 & 40.43 & 46.03 & 24.35 & 32.67 & 34.71 & 20.10  & 19.85 & 38.86 \\
          & +SAR  & 76.55 & 55.13 & 40.42 & 45.86 & 24.73 & 33.07 & 34.62 & 21.09 & 19.74 & 39.02 \\
          & Ours  & 76.55 & \textbf{55.23} & \textbf{41.02} & \textbf{46.56} & \textbf{25.94} & \textbf{34.44} & \textbf{35.23} & \textbf{21.81} & \textbf{20.35} & \textbf{39.68} \\
    \midrule
    \midrule
    \multirow{5}[2]{*}{miniImageNet-C} & +TENT & 75.37 & 12.74 & 50.78 & 41.59 & 19.07 & 39.45 & 38.39 & 22.71 & 22.64 & 35.86 \\
          & +SAM  & 75.37 & 13.21 & 52.79 & 41.76 & 20.41 & 41.72 & 39.02 & 23.22 & 23.07 & 36.73 \\          
          & +GSAM & 75.37 & \textbf{14.67} & 54.68 & 43.74 & 21.19 & 43.01 & 41.04 & 25.04 & \textbf{26.41} & 38.35 \\
          & +SAR  & 75.37 & 13.16 & 54.13 & 43.22 & 21.76 & 43.23 & 40.37 & 24.43 & 25.92 & 37.95 \\
          & Ours  & \textbf{75.37} & 13.51 & \textbf{55.71} & \textbf{44.03} & \textbf{22.93} & \textbf{44.94} & \textbf{42.78} & \textbf{25.65} & 26.04 & \textbf{39.00} \\
    \bottomrule
    \end{tabular}%
  \label{tab:Adaptation_Com_cifar}%
\end{table*}%

\begin{table*}[]
  \centering
  \caption{Comparison with TTA/DG baselines on CUB200-C dataset. The corruptions for sessions 1-10 are 'pixelate',  'elastic transform', 'defocus blur', 'zoom blur', 'glass blur', 'snow', 'fog', 'Gaussian noise', 'shot noise', 'impulse noise'.}
    \begin{tabular}{ccccccccccccc}
    \toprule
    \multirow{2}[4]{*}{Method} & \multicolumn{11}{c}{Accuracy in each session (\%)}                                    & \multirow{2}[4]{*}{Average Accuracy (\%)} \\
\cmidrule{2-12}          & 0     & 1     & 2     & 3     & 4     & 5     & 6     & 7     & 8     & 9     & 10    &  \\
    \midrule
    +TENT & 81.87 & 62.08 & 59.72 & 46.67 & 25.49 & 32.41 & 30.87 & 32.51 & 24.33 & 24.72 & 18.64 & 39.93 \\
    +SAM  & 81.87 & 63.19 & 61.49 & 48.22 & 26.98 & 33.79 & 32.19 & 34.10  & 25.32 & 25.79 & 18.87 & 41.07 \\
    +SAR  & 81.87 & 64.03 & 62.01 & \textbf{49.65} & 27.72 & 33.94 & 33.41 & 35.29 & 26.07 & 27.25 & 20.27 & 41.95 \\
    +GSAM & 81.87 & 64.76 & 61.72 & 48.64 & 27.57 & 34.07 & 33.21 & 35.45 & 25.97 & \textbf{27.62} & 19.42 & 41.84 \\
    \textbf{Ours} & 81.87 & \textbf{65.23} & \textbf{62.25} & 49.21 & \textbf{28.77} & \textbf{34.23} & \textbf{33.96} & \textbf{36.37} & \textbf{26.61} & 27.53 & \textbf{20.83} & \textbf{42.44} \\
    \bottomrule
    \end{tabular}%
  \label{tab:Adaptation_Com_CUB}%
\end{table*}%
\subsection{Overall Results}
\textbf{Comparison with CIL/FSL/FSCIL methods.}
To validate our method, we conducted comparisons with current approaches in CIL/FSL/FSCIL on three benchmarks, i.e., CIFAR100-C, CUB200-C, and miniImagenet-C. Table~\ref{tab:CUB200-C} shows all incremental and average accuracies on the CUB200-C dataset. Considering the average accuracy across all sessions, our method achieves an average accuracy of 42.44\%, surpassing the performance of the compared methods. This indicates that our approach exhibits superior performance and stability in handling generalized incremental learning under concept drift. Specifically, iCaRL and ProtoNet initially demonstrate relatively high performance in the base session (session 0). However, due to the challenge of label scarcity, these methods fail to acquire information about new classes. Moreover, when faced with the covariate shift between the source and target streams, the model performance further deteriorates, resulting in the poorest performance of all the methods. In contrast, few-shot class incremental learning methods such as TOPIC, CEC, FACT, and TEEN effectively mitigate the label scarcity issue, leading to improved performance. However, these methods still struggle to overcome distribution shift problems. Our proposed method addresses these challenges effectively. It not only learns new classes with limited labeled data incrementally but also makes adaptations from the labeled source stream to the unlabeled target stream. Consequently, our method achieves the best performance.

The same conclusion can also be observed in Table~\ref{tab:cifarAndImage}, which shows the overall accuracy on the CIFAR100-C and miniImagenet-C datasets. The results also show that our method consistently outperforms the others in performance. In particular, as new sessions come, our method demonstrates superior performance in subsequent sessions. This further demonstrates the superior performance of our method in handling class-incremental and distribution changes in dynamic streaming environments.

\noindent
\textbf{Comparison with TTA/DG methods.} To demonstrate the adaptation capability of the RSGS minimization algorithm,  we conducted comparisons with the TTA and DG methods on the CIFAR100-C and CUB200-C datasets. To ensure fair comparisons, we kept the training-free calibrated prototypes strategy fixed for the incremental learning of new classes and only employed different methods during the adaptation stage.

As shown in Table~\ref{tab:Adaptation_Com_cifar}, it is evident that our method consistently outperforms the TTA/DG baselines across all incremental sessions on the CIFAR100-C and miniImageNet-C datasets. Specifically, TENT performs the worst because TENT only minimizes the entropy, which can lead to overfitting issues during training and convergence towards sharp minima. Therefore, SAM minimizes the sharpness of the entropy, enhancing the model's generalization. SAR further removes samples with high entropy to improve model generalization. However, the perturbed loss is not always sharpness-aware, so GSAM minimizes the surrogate gap and perturbed loss simultaneously. Our RSGS method comprehensively addresses all these issues. It not only minimizes the perturbation loss and the surrogate gap simultaneously but also integrates a reliable indicator function to filter out samples with high entropy. Therefore, it can achieve the best results, highlighting its robustness and superiority in handling incremental learning tasks under distribution shifts. The same results can also be verified in Table~\ref{tab:Adaptation_Com_CUB}, which demonstrates its superior performance on the CUB200-C dataset.

\subsection{Ablation Study}
\begin{figure}[htbp]
    \centering
    \includegraphics[width=1\columnwidth]{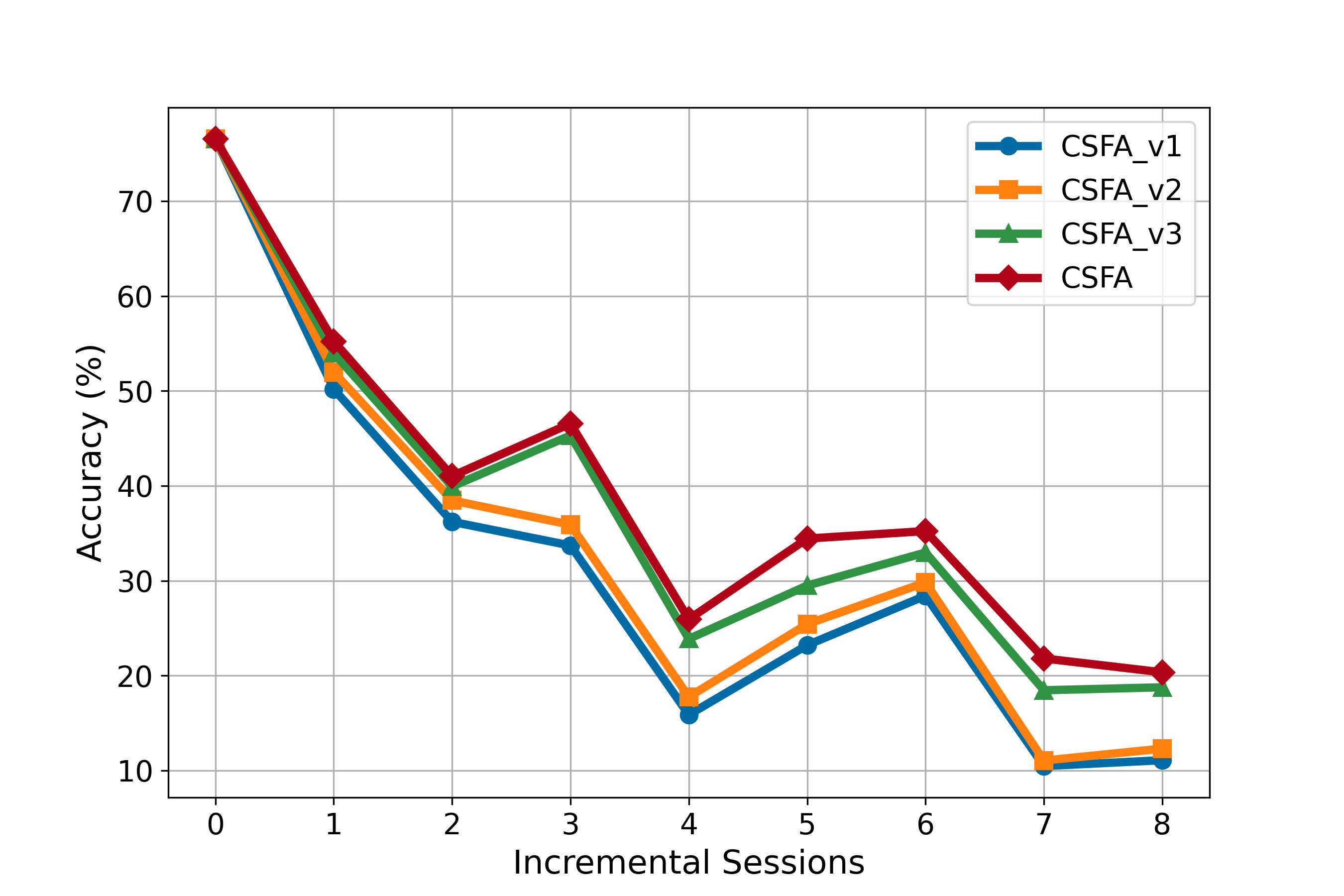}
    \caption{Incremental accuracy (\%) of CSFA variants.}
    \label{fig:ablation}
\end{figure}

To comprehensively evaluate the efficacy of each component in CSFA, we perform an ablation study on the CIFAR100-C dataset. Specifically, we devised three variations of CSFA to validate the rationale of each component and its impact on the overall classification results. 
As shown in Figure~\ref{fig:ablation}, CSFA$_{v1}$ serves as a baseline that does not consider limited samples and drift adaptation. This method solely employs incremental learning to fine-tune the model with new samples for predicting the target stream. Consequently, for the GILCD task, CSFA$_{v1}$ performs the poorest and exhibits significantly lower performance than CSFA. CSFA$_{v2}$ incorporates a prototype classifier without calibrations for few-shot class-incremental learning, leading to improved performance. However, due to the evident covariate shift between the source and target streams, its predictive results remain unsatisfactory. Furthermore, we introduce our proposed RGSM adaptation strategy into CSFA$_{v3}$, resulting in a significant improvement in predictive performance. Lastly, the complete CSFA further utilizes the well-learned base prototypes to calibrate the biased prototypes of novel classes during incremental sessions, achieving the best performance.

\subsection{Influence of Incremental Shot}
\begin{figure}[htbp]
    \centering
    \includegraphics[width=1\columnwidth]{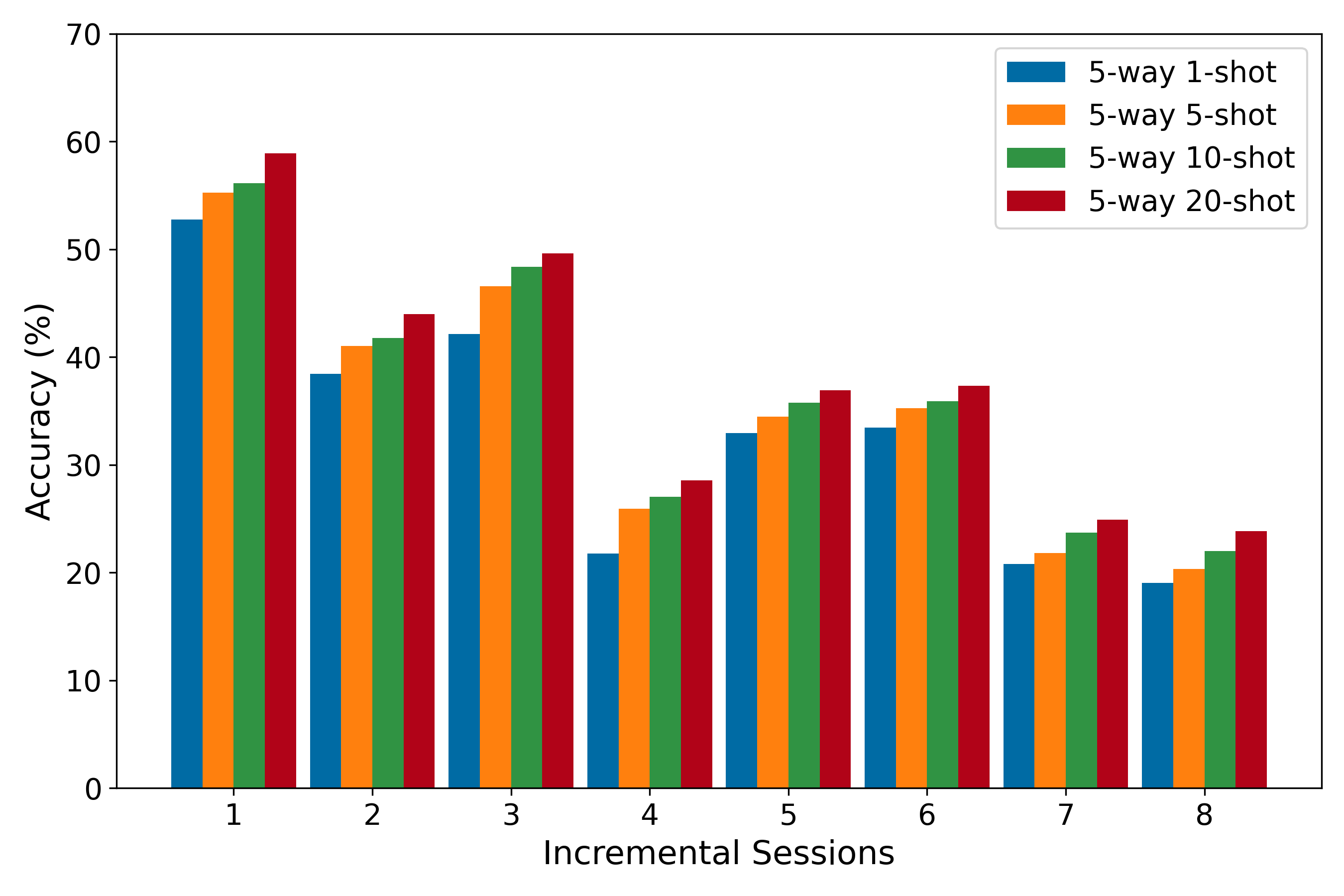}
    \caption{Influence of incremental shot.}
    \label{fig:increShots_CIFAR100}
\end{figure}

In the GILCD setting, we assume that each incoming session follows an $N$-way $K$-shot setup. Therefore, the incremental learning performance depends on the number of new samples provided in each session, i.e., $K$. Thus, we vary the shot number to investigate its impact on the final accuracy. We keep the incremental way consistent with the benchmark setting and change the shot number $K$ among \{1, 5, 10, 20\} on the CIFAR100-C dataset. As inferred from Figure~\ref{fig:increShots_CIFAR100}, with more instances per class in the source stream, the model can learn more information about new classes. Thus, the estimation of prototypes becomes more precise, leading to improved performance.

\section{Conclusion \& Limitations}
\label{conclusion}

In this work, we tackled the challenge of learning from evolving data streams characterized by concurrent concept drift and class evolution. We formalized this setting as GILCD, highlighting the compounded difficulties of scarce supervision and pervasive uncertainty. To address them, we proposed CSFA, a unified framework that combines a \textit{training-free calibrated prototype strategy} for efficient assimilation of novel classes with a \textit{source-free adaptation mechanism} (RSGS) for robust drift handling. By calibrating prototypes without retraining and filtering unreliable target samples during adaptation, CSFA ensures both semantic stability and distributional robustness. Extensive evaluations demonstrate its effectiveness under real-world streaming conditions, advancing adaptive learning toward more resilient and practical systems in non-stationary environments.

While CSFA shows strong performance, it still has limitations. CSFA relies on implicit adaptation without explicit drift detection, which may cause unnecessary updates. Incorporating detection could improve efficiency by triggering adaptation only under significant shifts. Moreover, GILCD assumes novel classes are strictly disjoint, overlooking practical scenarios such as open-set recognition and long-tail evolution that warrant further exploration.

\begin{acks}
The work was supported by the Australian Research Council under Laureate project FL190100149 and discovery project DP220102635.
\end{acks}

\newpage
\bibliographystyle{ACM-Reference-Format}
\balance
\bibliography{ref}

\appendix
\section{Related Work}
\subsection{Concept Drift}
The realm of data stream learning has attracted significant research attention due to the dynamic attributes of real-world streaming data, i.e., concept drift. Concept drift refers to the phenomenon of a shift in data distributions over time, which occurs when the joint distribution $P_{t+1}(X, y)$ at time $t+1$ differs from $P_t(X, y)$ at time $t$~\cite{lu2018learning,yu2024gnn,song2021learning,yang2025walking}. It poses formidable challenges to maintaining classifier accuracy and ensuring rapid adaptability. To tackle this issue, various methods for concept drift adaptation have been developed to enhance model effectiveness and reliability in the presence of concept drift, such as window-based methods~\cite{xu2017dynamic}, instance-based approaches~\cite{gomes2017adaptive}, ensemble learning-based algorithms~\cite{wang2016online} and so on. Concept drift adaptation ensures that predictive models can adjust to newly arriving data within evolving distribution environments. In addition, these models also exhibit robustness to noise while minimizing memory and time costs~\cite{celik2021adaptation,wen2024adaptive}.

While many existing methods are designed for single-labeled streams, adaptive learning for multiple streams under concept drift presents a more complex and challenging scenario~\cite{yu2024online,duan2025bayesian,yu2024fuzzy}. In recent years, the field of adaptive learning for multiple streams has garnered significant attention. For example, a prominent framework for multistream classification~\cite{chandra2016adaptive} defines a labeled source stream generated in a non-stationary process and an unlabeled target stream from another dynamic process. This framework aims to predict the class labels of target instances using label information from the source stream. The FUSION algorithm~\cite{haque2017fusion} further improves this approach with effective density ratio estimation. Furthermore, neural network-based models, such as Autonomous Transfer Learning~\cite{pratama2019atl}, are developed to handle high-dimensional data, employing generative and discriminative phases combined with Kullback-Leibler divergence-based optimization. Additionally, a meta-learning-based framework~\cite{yu2022meta} is proposed to learn invariant features of drifting data streams and update the meta-model online.

However, these studies have overlooked a critical issue evident in real-world applications: learning must occur continuously on incoming data streams, which may include data from both new classes and existing classes. Despite the advancements made in concept drift adaptation methods for multistream classification, they still do not adequately address the problem of class incremental learning in dynamic environments.

\subsection{Class-Incremental Learning}
CIL endeavors to construct a comprehensive classifier encompassing all encountered classes over time~\cite{masana2022class}. The primary challenge in CIL, known as catastrophic forgetting, arises when optimizing the network with new classes leads to the loss of knowledge about previous classes, resulting in irreversible performance degradation. Thus, to effectively mitigate the catastrophic forgetting problem, there are three major categories: replay-based methods~\cite{gu2022not}, knowledge distillation~\cite{rebuffi2017icarl}, and model expansion~\cite{wang2024few}. Replay-based approaches offer an intuitive means of leveraging previous data for rehearsal, enabling the model to revisit former classes and resist forgetting. Alternatively, some methodologies incorporate regularization terms with additional data to guide optimization direction and mitigate catastrophic forgetting. Knowledge distillation-based CIL methods aim to establish mappings between old and new models, thereby preserving the characteristics of the old model during the updating process. Additionally, recent studies have demonstrated the effectiveness of model expansion in CIL~\cite{yan2021dynamically}. A notable method involves preserving a single backbone and freezing it for each incremental task, effectively alleviating the catastrophic forgetting problem.

CIL methods often depend on abundant labeled data for supervised learning. This presents a significant challenge, especially in real-world data streams with limited runtime and labeled data for model updates. In response to this practical concern, FSCIL is introduced, which aims to efficiently tackle the class-incremental learning problem with only limited labeled data available. This paradigm has demonstrated effectiveness in addressing the challenges of class-incremental learning under resource constraints~\cite{zhang2021few}. For instance, Tao et al.~\cite{tao2020few} proposed a neural gas network to preserve the topology of features in both the base and new classes for the FSCIL task. In~\cite{zhang2021few}, they introduced a continually evolved classifier for few-shot incremental learning, utilizing an adaptation module to update classifier weights based on a global context of all sessions. Furthermore, Wang et al.~\cite{wang2024few} proposed a training-free prototype calibration strategy (TEEN) to calibrate the biased prototypes of new classes. However, all these methods are designed for static scenarios and do not address the problem of distribution changes.

\begin{figure*}[th]
    \centering
    \subfigure[$\tau$]{
        \includegraphics[width=0.4\textwidth]{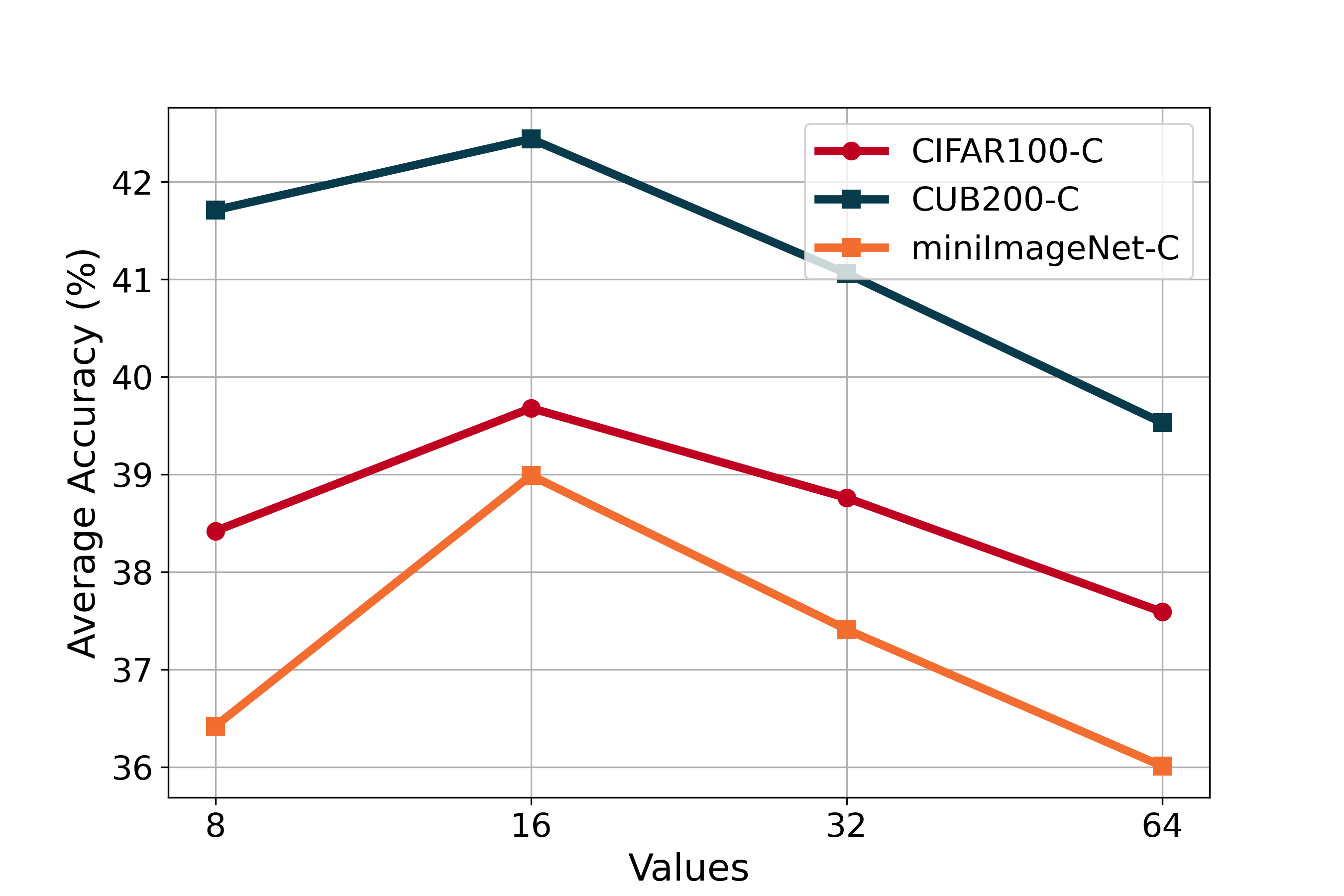}
        \label{fig:tao}
    } 
    \subfigure[$\alpha$]{
        \includegraphics[width=0.4\textwidth]{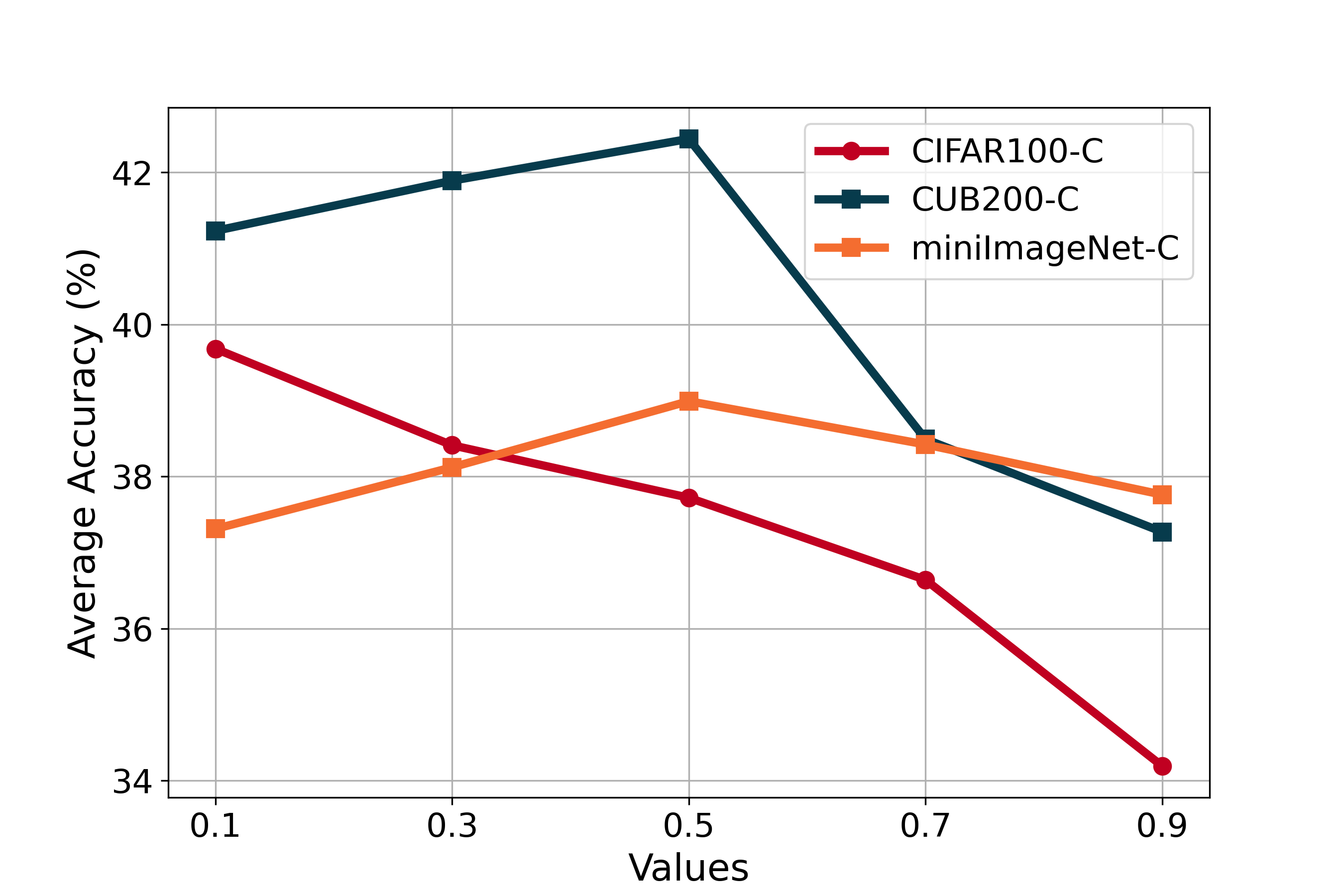}
        \label{fig:alpha}
    }
   \subfigure[$\beta$]{
        \includegraphics[width=0.4\textwidth]{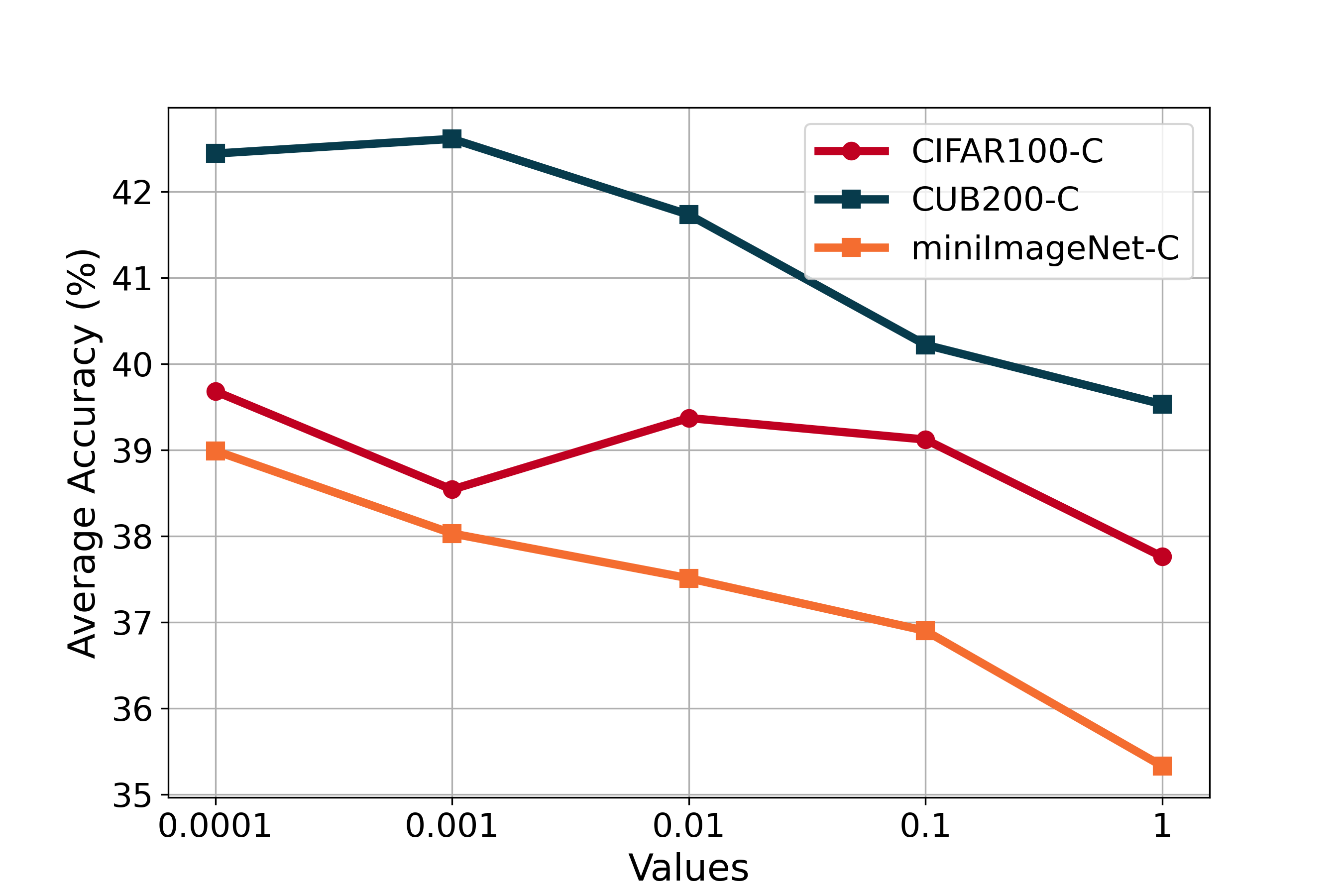}
        \label{fig:beta}
    } 
    \subfigure[Batch size]{
        \includegraphics[width=0.4\textwidth]{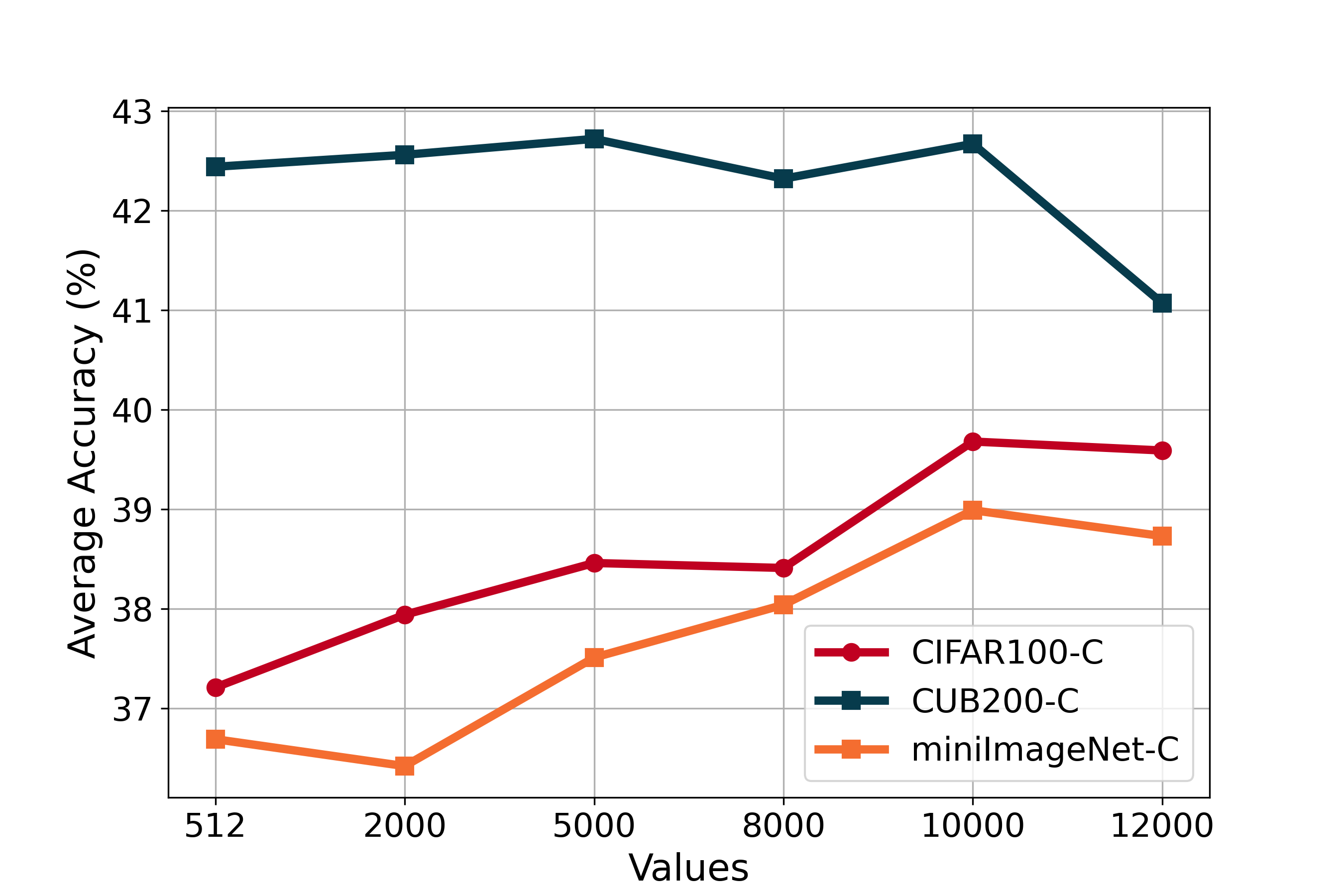}
        \label{fig:batchsize}
    } 
    \caption{The effect of different parameters on average classification accuracy.}
    \label{fig:para}
\end{figure*}

\subsection{Test-Time Adaptation and Domain Generalization}
Test-Time Adaptation (TTA) has garnered significant attention in recent years due to its potential to enhance model performance in real-world non-stationary scenarios~\cite{liang2023comprehensive}. Similarly, source-free domain adaptation also requires no access to the training (source) data~\cite{liang2020we}. This technique involves adapting a trained model during inference time to better suit the characteristics of the current data distribution. Various approaches have been proposed to achieve Test-Time Adaptation across different domains. Current TTA methods can be categorized as test-time training and fully test-time adaptation based on whether they alter the training process. Test-time training methods jointly optimize a source model with both supervised and self-supervised losses, such as contrastive-based objectives, rotation prediction and so on. It then conducts self-supervised learning at test time to fit the new test distributions~\cite{sun2020test}. Fully test-time adaptation methods do not modify the training process and can be applied to any pre-trained model. These techniques involve adapting various aspects such as statistics in batch normalization layers~\cite{wang2020tent}, entropy minimization~\cite{niu2022towards}, prediction consistency maximization~\cite{chen2022contrastive}, top-$k$ classification boosting~\cite{niu2022boost} and so on.

Domain Generalization aims to develop a model from one or several observed source datasets that can effectively generalize to unseen target domains~\cite{wang2022generalizing,yu2025learning}. Current domain generalization methods approach domain generalization from various perspectives, including invariant representation, data augmentation, and empirical risk minimization. In this paper, we specifically concentrate on empirical risk minimization methods to address domain generalization challenges. For example, SAM~\cite{foret2020sharpness} is proposed for domain generalization, enhancing generalization by minimizing both loss value and sharpness simultaneously. GSAM~\cite{zhuang2022surrogate} introduces a surrogate gap to quantify the difference between the maximum loss within the neighborhood and the minimum point. This approach offers a more precise characterization of sharpness and enhances generalization by simultaneously minimizing sharpness-aware entropy and the surrogate gap.

\section{Experiments}

\subsection{Baselines}
\label{app:baselines}
To validate our approach, we commence by comparing it with the state-of-the-art methods of CIL, FSL, and FSCIL:
\begin{itemize}
    \item iCaRL~\cite{rebuffi2017icarl} stands out as a prominent approach for class-incremental learning.  It adopts nearest-mean-of-exemplars classifiers to store exemplars from each class and utilizes a herding-based strategy for exemplar selection. By integrating these exemplars with distillation, it effectively learns data representations to overcome catastrophic forgetting.
    
    \item ProtoNet~\cite{snell2017prototypical} is a widely recognized method in the field of few-shot learning. It constructs a metric space where classification is executed by measuring distances to prototype representations of individual classes. 
    
    \item TOPIC~\cite{tao2020few} introduces a neural gas network designed to grasp and maintain feature topology across diverse classes. It can effectively prevent the forgetting problem and enhance representation learning for new classes by dynamically adapting the network to new samples.
    
    \item CEC~\cite{zhang2021few} proposes a continually evolved classifier for few-shot class-incremental learning. It incorporates a graph attention network to dynamically update classifier weights, leveraging a global context across all sessions.
    
    \item FACT~\cite{zhou2022forward} emphasizes the necessity of constructing forward-compatible models for few-shot class incremental learning. By pre-assigning virtual prototypes in an embedding space, the model becomes anticipatory and expandable, effectively mitigating the impact of model updates and improving inference performance.
    
    \item TEEN~\cite{wang2024few} addresses the issue of biased prototypes for new classes by calibrating them with well-calibrated prototypes from old classes. Also, this method eliminates the need for additional optimization procedures.
\end{itemize}

In addition, we also compare our proposed adaptation method with current TTA and Domain Generalization (DG)  methods to demonstrate its adaptability to different distributions.
\begin{itemize}
    \item TENT~\cite{wang2020tent} is a fully test-time adaptation method that reduces generalization error on shifted data through entropy minimization without accessing the source data.
     \item SAM~\cite{foret2020sharpness} is proposed for domain generalization, enhancing generalization by minimizing both loss value and sharpness simultaneously.
    \item GSAM~\cite{zhuang2022surrogate} introduces a surrogate gap to quantify the difference between the maximum loss within the neighborhood and the minimum point. This approach offers a more precise characterization of sharpness and enhances generalization by simultaneously minimizing sharpness-aware entropy and the surrogate gap.
    \item SAR~\cite{niu2022towards} is a more stable and reliable fully test-time adaptation method, which mitigates the impact of noisy test samples with large gradients.
\end{itemize}

\subsection{Implementation}
\label{app:imple}
In the experiment, we adopt the ResNet20 for CIFAR100-C, a pre-trained ResNet18 for CUB200-C, and a randomly initialized ResNet18 for miniImageNet-C. To ensure a fair comparison, all methods utilize identical backbone networks and initialization protocols. For training the feature extractor on CIFAR100-C and miniImageNet-C, we employ a learning rate of 0.1 and a batch size of 256. Furthermore, we adjust the learning rate to 0.004, set the batch size to 128, and conduct training over 400 epochs on CUB200-C. To manage learning rate adjustments, we implement a cosine scheduler. For our RSGS adaptation process, we use SGD to update the model with a momentum of 0.9, and the learning rate is 0.0001. All experiments are conducted using PyTorch on an A100 GPU.


\subsection{Parameter sensitivity.}
There are four main parameters affecting the performance, i.e., the scaling hyperparameter $\tau$, calibration coefficient $\alpha$, ascent step size $\beta$, and the batch size during adaptation. To analyze their impact on the overall performance, we conduct experiments under various values of all parameters on these three datasets. Specifically, we search the optimal parameters by setting $\tau \in \{8, 16, 32, 64\}$, $\alpha \in \{0.1, 0.3, 0.5, 0.7, 0.9\}$, $\beta \in \{0.0001, 0.001, 0.01, 0.1, 1\}$ and  $batch\_size \in \{512, 2000, 5000, 8000, 10000, 12000\}$. Each parameter is tuned while the others are kept fixed, and the results are shown in Figure~\ref{fig:para}. 

During the class incremental learning, as our method does not require additional model updates after the base training, we need to determine the optimal performance by adjusting the scaling hyperparameter $\tau$ and calibration coefficient $\alpha$. 
Observations from Figure~\ref{fig:tao} and Figure~\ref{fig:alpha} reveal that various values of $\tau$ and $\alpha$ yield distinct outcomes. In our experiment, we set $\tau$ to 16 across all datasets, while assigning $\alpha$ values of 0.1 for CIFAR100-C and 0.5 for CUB200-C and miniImageNet-C datasets. During the adaptation process, the parameter $\beta$ dictates the ascent step size. Figure~\ref{fig:beta} illustrates the varied effects of different settings of $\beta$ on model adaptation. Therefore, we opt for $\beta = 0.0001$ for CIFAR100-C and miniImageNet-C datasets, and $\beta=0.001$ for the CUB200-C dataset. Additionally, as depicted in Figure~\ref{fig:batchsize}, the choice of batch size significantly influences the adaptation process. Smaller batch sizes may introduce randomness, thereby yielding unstable results. Through meticulous experimental analysis, we designate the adaptation batch sizes to be 512 for CUB200-C and 10000 for CIFAR100-C and miniImageNet-C to achieve optimal performance.

\end{document}